\documentclass[twoside]{article}

\usepackage{aistats2021}
\usepackage{caption}
\usepackage{amsmath,amssymb}
\usepackage{graphicx}
\usepackage{subcaption}
\usepackage{xspace}
\usepackage{pgfplots}
\usepackage{balance}
\usepackage{microtype}
\usepackage{multirow}
\usepackage[hidelinks]{hyperref}
\usepackage{float}

\usepackage{enumitem}
\usepackage[noabbrev,capitalize]{cleveref}

\newcommand{\Remark}[1]{\textcolor{blue}{\sf [#1]}}

\newcommand{\ee}{\varepsilon}
\newcommand{\Rsp}{\mathbb{R}}
\newcommand{\knn}{{$k$-NN}\xspace}
\newcommand{\ccomp}{\mathcal{C}} 
\newcommand{\btdist}{d_B} 
\newcommand{\conf}{\phi}  
\newcommand{\dgm}{\operatorname{Dgm}}
\newcommand{\pers}{\operatorname{pers}} 
\newcommand{\ft}{f_\theta}
\newcommand{\best}[1]{\textbf{#1}}
\newcommand{\loss}{{\cal L}}
\newcommand{\reg}[1]{\ell_{#1}}

\usepackage{titlesec}
\titlespacing{\paragraph}{%
  0pt}{
  0.2\baselineskip}{
  .5em}
\titlespacing*{\section}{0pt}{2ex}{1ex}

\begin{document}
\twocolumn[

\aistatstitle{Topological Regularization via Persistence-Sensitive Optimization}

\aistatsauthor{Arnur Nigmetov$^{*}$ \And Aditi S. Krishnapriyan$^{*}$ \And Nicole Sanderson \And Dmitriy Morozov}

\aistatsaddress{ Computational Research Division, Lawrence Berkeley National Laboratory, Berkeley, CA 94720 }

]

\begin{abstract}
    Optimization, a key tool in machine learning and statistics, relies on
    regularization to reduce overfitting. Traditional regularization methods
    control a norm of the solution to ensure its smoothness. Recently,
    topological methods have emerged as a way to provide a more precise and
    expressive control over the solution, relying on persistent homology
    to quantify and reduce its roughness.  All such existing
    techniques back-propagate gradients through the persistence diagram, which is
    a summary of the topological features of a function. Their downside is that
    they provide information only at the critical points of the function. We
    propose a method that instead builds on persistence-sensitive simplification
    and translates the required changes to the persistence diagram into changes
    on large subsets of the domain, including both critical and regular points.
    This approach enables a faster and more precise topological regularization,
    the benefits of which we illustrate with experimental evidence.
\end{abstract}

\section{Introduction}
\label{sec:introduction}

Regularization is key to many practical optimization techniques. It allows the
user to add a prior about the expected solution --- e.g., that it needs to be
smooth or sparse --- and optimize it together with the main objective function.
Classical regularization techniques~\cite{convex-optimization}, such as $\reg{1}$- and $\reg{2}$-norm
regularization, have been studied in statistics and signal processing since at
least the 1970s. These techniques are especially important in machine learning,
where problems are often ill-posed and regularization helps prevent overfitting.
Accordingly, various regularization techniques are not only used in
machine learning research~\cite{learning-with-kernels,Ng2004}, but are also
incorporated into the standard optimization software and routinely used in
applications.

Recently, several authors have begun to explore the use of topological
methods to regularize the objective function. All of them use persistent
homology to measure either the shape of the data set or the topological
complexity of the learned function. For instance, Chen et
al.~\cite{topological-regularizer} use persistence to describe the complexity of
the decision boundary in a classifier and add terms to the loss to keep this
boundary topologically simple. Br\"uel-Gabrielsson et al.~\cite{topology-layer}
use persistence as a descriptor of the topology of the data and introduce a
family of losses to control the shape of the data once it passes through a neural
network.

All the methods that incorporate persistence into the loss function
\cite{topological-regularizer, topology-layer, differential-calculus-barcodes}
rely on the same observation. Persistent
homology describes data via a diagram, a collection of points $\{b_i,d_i\}$ in
the plane, that encodes the topological features of the data: components of the
decision boundary, ``wrinkles'' in the learned function, cycles in the point
set once it passes through the neural network. Each point represents the birth
$b_i$ and death $d_i$ of a topological feature. Each
coordinate depends on the value of the function on a set of points. In the
simplest case, $(b_i,d_i) = (f(x), f(y))$ for some $x,y$ in the input, where $f$
is the learned function. In the
more sophisticated cases, each point in the persistence diagram is generated by
a handful of input points (e.g., four~\cite{topology-layer}).
Accordingly, if a loss $\loss$ prescribes moving a point in the persistence diagram
via a gradient $(\partial \loss / \partial b_i, \partial \loss / \partial d_i)$,
one can back-propagate it
to update the model parameters.

\begin{figure*}
    \centering
    \includegraphics[width=\textwidth]{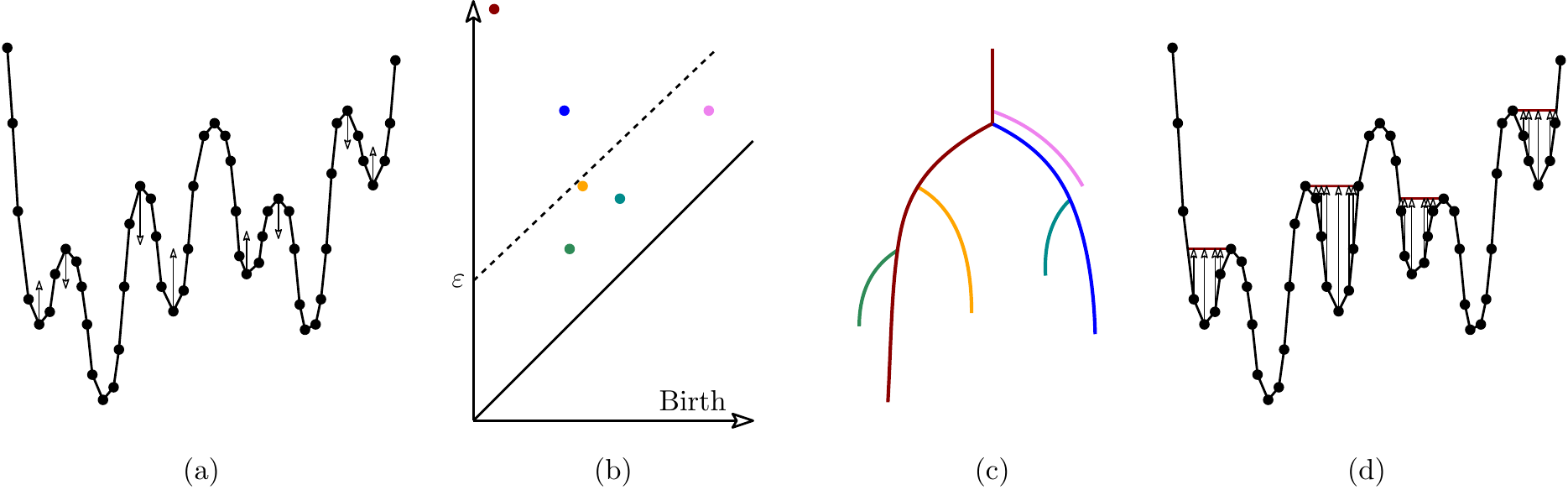}
    \caption{(a) Function on a graph, with gradients on critical points
                 prescribed by the diagram loss.
             (b) Persistence diagram of this function. Points closer to the
                 diagonal correspond to smaller fluctuations in the function,
                 and we interpret them as topological noise. $\ee$ indicates the
                 level of desired simplification that generates the gradients in
                 (a) and (d).
             (c) Merge tree of the function, with branches highlighted in
                 different color. The branches translate into the points in the
                 persistence diagram of the matching color.
             (d) Gradients prescribed by the persistence-sensitive optimization
                 (PSO loss). The gradients are present both on critical and
                 regular points.}
    \label{fig:simplification}
\end{figure*}

Although persistent homology describes a family of topological features of
different dimensions (connected components, loops, voids), most practical
examples have focused on 0-dimensional features (connected components generated
by the extrema of the input function). In this case, a natural loss is one that
penalizes and tries to remove low-persistence features, which are interpreted as noise: e.g.,
$\loss(f) = \sum_{(d_i - b_i) \leq \ee} (d_i-b_i)^2$.
\emph{Persistence-sensitive simplification}~\cite{simplification-2manifolds,simplification-linear-time,Bauer2012}
offers a direct solution to this problem. It prescribes how to modify a given input function $f$ to find a function $g$ that
is $\ee$-close to $f$, but without the noisy features. Given such a $g$,
which by construction minimizes the diagram loss $\loss$ above, one can use $\|f - g\|^2$
as a term in the loss. In the context of learning, this approach
offers a major advantage: instead of supplying gradients only on the critical
points of $f$, we also get gradients on the regular points of $f$ whose
values must be changed to topologically simplify the function; see
\cref{fig:simplification}.

Our contributions are:
\begin{itemize}[noitemsep,nolistsep]
    \item
        a method to control the topological complexity of a function,
        represented by a neural network, by incorporating persistence-sensitive
        simplification into the training;
    \item
        comparison of the training results after back-propagating gradients
        through the diagram vs.\ using persistence-sensitive optimization;
    \item
        experiments with data that illustrate the utility of controlling the
        topology of the learned function.
\end{itemize}

We note that topological methods have found a much broader use in machine
learning than regularization. An important line of work involves developing
techniques to incorporate topological features detected in data into machine
learning algorithms~\cite{HKN2019, perslay, persistence-images}. Although there
is some overlap in methods between the two research directions (notably propagating
loss through the persistence diagram), our work is focused on regularization.

\section{Background}
\label{sec:background}

We recall the relevant background in topological data
analysis~\cite{comp-top-book}, focusing
specifically on 0-dimensional persistent homology, which we introduce using an
auxiliary computational construction, merge trees.

\paragraph{Merge trees.}
Let $f \colon X \to \Rsp$ be a function on a topological space $X$.
A \emph{merge tree} tracks evolution of connected components in the sub-level
sets $f^{-1}(-\infty, a]$ of the function, as we vary the threshold $a$.
Formally, we identify two points $x$, $y$
of $X$, if $f(x) = f(y) = a$ and $x$ and $y$ belong to the same connected
component of the sub-level set $f^{-1}(-\infty, a]$.
The quotient of $X$ by this equivalence relation
is called a merge tree of $f$.

Throughout the paper we use graphs to approximate continuous spaces, so we
briefly dissect the above definition for functions on graphs.
Let $f \colon G \to \Rsp$ be a function on a graph $G = (V, E)$, defined on the
vertices and linearly interpolated on the edges.
For simplicity, we assume that all the values of $f$ on
the vertices are distinct and
index the vertices $V = \{ v_i \}$  so that $f(v_i) < f(v_{i+1})$.
The \emph{merge tree} of $f$ is a graph $T = (V, E_T)$
such that an edge $(v_i, v_j)$ for $i < j$ is present in $T$ if and only if
$v_i$ and $v_j$ belong to the same connected component $\ccomp$
of $f^{-1}(-\infty, f(v_j)]$ and there does not exist $k$ such that $i < k < j$ and $v_k \in \ccomp$.
A merge tree $T$ is not necessarily a tree --- it is a forest, with a tree for
every connected component of $G$ --- but the distinction is minor for this
paper.

$T$ is naturally decomposed into \emph{branches}; see \cref{fig:simplification}.
A branch $B \subseteq V$ tracks a component of the sub-level
set of $f$ that first appears at a local minimum $v_b \in B$.
This component disappears by merging into another branch $B'$ that appeared at a
lower local minimum $v'_b$. $B$ merges into $B'$ at a saddle $v_d \in B'$.
We say that $B$ is born at $f(v_b)$ and it dies at $f(v_d)$.
The branch of the tree, born at the global minimum, that never merges into a
deeper branch dies at $\infty$, by definition. 
The \emph{persistence} $\pers(B)$ of a branch $B$ is defined as the absolute
value of the difference between its death and birth values.


\paragraph{Persistence.}
A 0-dimensional persistence diagram, denoted $\dgm(f)$, is another
summary of the connectivity of the sub-level sets of $f$.
It is a multiset of points in the (extended) plane:
a branch $B$, born at $f(v_b)$ that dies at $f(v_d)$ is summarized by the point
$(f(v_b),f(v_d))$.
Points closer to the diagonal represent shorter branches and we interpret
them as noise.

Although we have defined everything in terms of the sub-level sets, the
definition for super-level sets, $f^{-1}[a,\infty)$ is symmetric, with maxima
replacing the minima. We use both constructions throughout the paper.

If graph $G$ has $n$ vertices and $m$ edges, then a merge tree on $G$
can be computed in $O(n \log n + m \alpha(m))$, where $\alpha$
is the inverse Ackermann function.
It follows that a 0-dimensional persistence diagram can be computed
in the same time. 

To visualize the topological changes in the model during optimization,
we stack persistence diagrams next to each other.
The resulting \emph{vineyard} of a family of functions
$f_i$ is a multiset of points $(i, | d^i_j - b^i_j|)$,
where $\{ (b^i_j, d^i_j) \}$ is the persistence diagram of $f_i$.
In other words, over each $i$ (for example, a training epoch) we plot all
persistences of the corresponding diagram.

\paragraph{Simplification.}
An important property of persistence is stability: a small perturbation of
function $f$ causes a small perturbation of the persistence diagram $\dgm(f)$.
The formal statement is the celebrated Stability Theorem:
\[
    \btdist(\dgm(f), \dgm(g)) \leq \| f - g \|_{\infty},
\]
where $f$ and $g$ are two real-valued functions  on the same domain and
$\btdist$ denotes the \emph{bottleneck distance}. This theorem is one of the
justifications for treating points close to the diagonal as topological noise.

This view suggests getting rid of the topological noise.
Let $f \colon G \to \Rsp$ be a function on a graph $G$.
A function $g \colon G \to \Rsp$ is called its \emph{$\ee$-simplification},
if $\| f - g \|_{\infty} \leq \ee$ and
$\dgm(g) = \{ (b,d) \in \dgm(f) \mid |d-b| > \ee \}$. In other words, $g$ is
$\ee$-close to $f$ but its persistence diagram has only those points whose
persistence exceeds $\ee$.
In the case of 0-dimensional persistence, $\ee$-simplification always exists and
can be computed in the same time as a merge
tree~\cite{simplification-2manifolds,simplification-linear-time,Bauer2012}.

\section{Method}
\label{sec:method}


We start with the standard supervised learning problem. Given training
data $x_i$ with labels $y_i$, we want to learn a model $\ft$, with
parameters $\theta$, that approximates $y_i$ given $x_i$. Although this
framework applies more generally, throughout the paper we focus on the case
where $\ft$ is a neural network.

Suppose we are solving a regression problem. In this case, the input labels are
scalars, $y_i \in \Rsp$, and our network maps from some (typically) Euclidean
space into reals, $\ft \colon \Rsp^d \to \Rsp$. The learning process is
usually a form of gradient descent on the network parameters with respect to a
user-chosen loss, for example, the mean-squared error (MSE),
$\loss(\theta) = \sum (\ft(x_i) - y_i)^2 / n$.

Ideally, we would like to topologically simplify the model $\ft$ either on its
entire domain, or at least on the ``data manifold,'' the subset of the domain
that contains all possible data. Unfortunately, there are no algorithms to solve
this problem --- topological methods require a combinatorial representation of
the domain --- so we resort to a standard approximation.


We take the domain of the network $\ft$ to be the $k$-nearest neighbors graph
on the training set $\hat{X}$: each training sample is a vertex, and two
vertices are connected if and only if one of them is among the $k$-nearest
neighbors of the other one.
The \knn graph $G$ approximates the data manifold. We can increase the quality
of this approximation by sampling additional points in the neighborhood of our
input. In the experiments in \cref{sec:experiments}, we draw $n$ additional
points from a normal distribution, centered on each training data point,
$x \in \hat{X}$, which results in a graph with $(n + 1) \cdot | \hat{X} |$ vertices.
(Although we don't know the true label on the extra points, we don't need it for
the topological simplification.)
Both because computing a \knn graph is expensive for high-dimensional data and
because it helps to control noise, in some experiments we build the \knn graph
on the lower-dimensional projection of $\hat{X}$ using PCA.

We use merge trees to compute an $\ee$-simplification $g$ of our model $\ft$.
For every vertex $v$, we find its first ancestor $u$ that lies on a
branch with persistence at least $\ee$. (If $v$ is already on such a branch,
then $u = v$.) We set $g(v) = \ft(u)$.  The effect of this operation on the
merge tree is that all the branches with persistence less than $\ee$ are
removed; see \cref{fig:simplification}.


\paragraph{Applying simplification.}
%
Given an $\ee$-sim\-pli\-fi\-ca\-ti\-on $g$ of $\ft$, we could add a term $\lambda \cdot \|\ft - g\|^2$
to the loss and use a single optimizer.
Instead, we opted for a different approach by alternating between the standard training
and the topological phases, with a separate optimizer for each phase.
A key advantage of this separation is that it keeps two histories of the
gradients, one for each phase, so that the topological loss does not influence
the momentum in the standard training.

An important decision is when to switch to the topological phase. We use a
heuristic that depends on the validation loss.
In each epoch, we first iterate over all batches and perform standard training
using the first optimizer. Then, if the validation loss increases, compared to
the previous epoch, by more than some threshold (a hyperparameter),
we compute the $\ee$-simplification $g$ and take $5$ to $10$ steps
with the second optimizer to minimize $\|\ft - g \|^2$.
We use the norms of the gradients of the ordinary training loss and
of the topological loss, to set a learning rate for the latter that ensures that
we update the model parameters $\theta$ by comparable amounts in both phases.

\paragraph{Choice of $\ee$.}
A key decision in implementing our method is how to choose $\ee$, to decide
which points to keep and which to remove in the persistence diagram.
Earlier works~\cite{topological-regularizer, topology-layer} prescribe a fixed
number of points to keep in a certain region of the persistence diagram. For
instance, some of the losses in \cite{topology-layer} penalize all but $j$ of
the most persistent points. We can optimize such a loss by
setting $\ee = (p_j + p_{j+1}) / 2$, where $p_i$ is the persistence of each
point, sorted in descending order.

Another alternative, used in topological data analysis to automatically
distinguish between persistent and noisy points, is the \emph{largest-gap
heuristic}. To apply it, we find index $j$ such that the difference
$p_j - p_{j+1}$ is maximized.

Finally, the heuristic that we found most effective and use for all experiments
in \cref{sec:experiments} is to use validation loss as our $\ee$.
Validation loss tells us how far we are from a function that gives perfect
answers on the validation set. Using it as $\ee$, we find the topologically simplest
function $g$ that is within the same distance from our model $\ft$.

\begin{figure*}[t]
    \begin{subfigure}[t]{.19\textwidth}
       \centering
        \includegraphics[width=\textwidth]{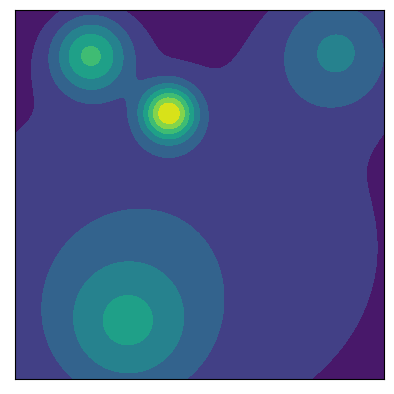}
        \caption{}
        \label{fig:values-original}
    \end{subfigure}
    \hfill
    \begin{subfigure}[t]{.19\textwidth}
        \centering
        \begin{tikzpicture}[scale=0.48]
            \begin{axis}[xlabel=step,
                         ytick style={draw=none},
                         yticklabels={,,},
                         ylabel={},
                         ylabel shift=1 mm]
                \pgfplotstableread[col sep=semicolon]{data/vineyard-values-pso.txt}\psodata
            \addplot[
                 scatter src=explicit symbolic,
                 scatter/classes={r={mark=*,red}, b={mark=*,black}, g={mark=*,green!60!black} },
                 scatter, only marks,
                 ]
                 table[x=x,y=y,meta=label] {\psodata};
            \addplot[mark=none, black, dashed] table[x=x, y=epsilon] {\psodata};
            \end{axis}
        \end{tikzpicture}
        \caption{}
        \label{fig:vineyard-values-pso}
    \end{subfigure}
    \hfill
    \begin{subfigure}[t]{.19\textwidth}
       \centering
       \includegraphics[width=\textwidth]{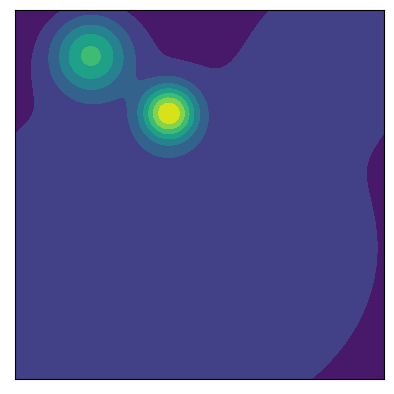}
        \caption{}
        \label{fig:simplified-values-pso}
    \end{subfigure}
    \hfill
    \begin{subfigure}[t]{.19\textwidth}
        \centering
        \begin{tikzpicture}[scale=0.48]
            \begin{axis}[xlabel=step,
                         ytick style={draw=none},
                         yticklabels={,,}
                         ylabel={},
                         ylabel shift=1 mm]
                \pgfplotstableread[col sep=semicolon]{data/vineyard-values-dgm.txt}\dgmdata
            \addplot[
                 scatter src=explicit symbolic,
                 scatter/classes={r={mark=*,red}, b={mark=*,black}, g={mark=*,green!60!black} },
                 scatter, only marks,
                 ]
                 table[x=x,y=y,meta=label] {\dgmdata};
            \addplot[mark=none, black, dashed] table[x=x, y=epsilon] {\dgmdata};
            \end{axis}
        \end{tikzpicture}
        \caption{}
        \label{fig:vineyard-values-dgm}
    \end{subfigure}
    \hfill
    \begin{subfigure}[t]{.19\textwidth}
       \centering
       \includegraphics[width=\textwidth]{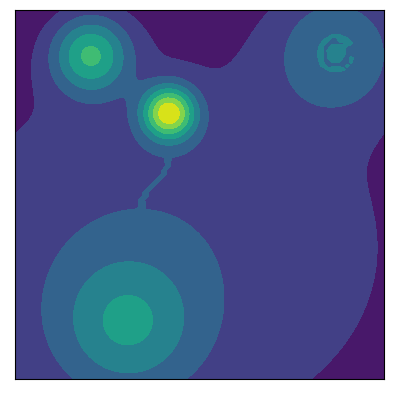}
        \caption{}
        \label{fig:simplified-values-dgm}
    \end{subfigure}
    \caption{Optimization of the values.
             (a) Original function.
             (b) Vineyard of simplification with PSO loss.
             (c) Function simplified with PSO loss.
             (d) Vineyard of simplification with diagram loss.
             (e) Function simplified with diagram loss.}
    \label{fig:dgm-vs-pso-values}
\end{figure*}
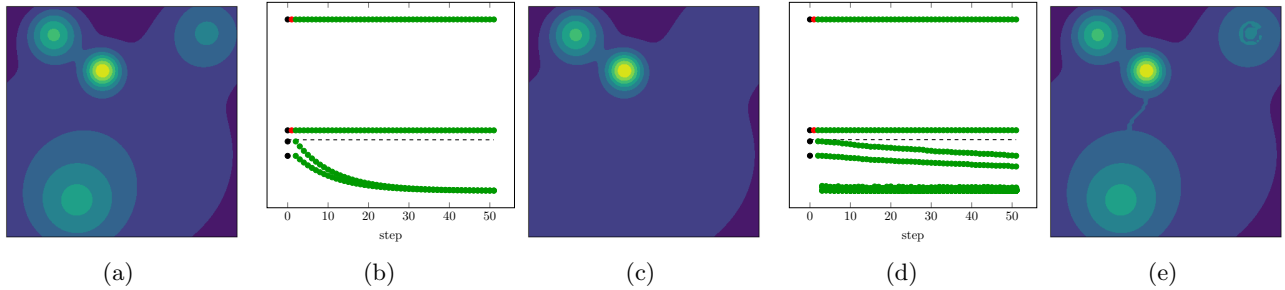

\paragraph{Classification.}
%
For regression, the network itself serves as a real-valued function amenable to
topological analysis.
Classification requires a little more work. We assume that the data has $m$ classes and the
network has $m$ output channels, $\ft \colon \Rsp^d \to \Rsp^m$, with the
predicted class chosen as $p = \arg\max_i \ft(x)[i]$.
We define the \emph{confidence function}, $\conf: \Rsp^d \to \Rsp$,
to measure how much higher the value in the predicted channel is compared to
the second highest candidate:
\begin{align*}
    \conf(x) & = \ft(x)[p] - \max_{i \neq p} \ft(x)[i].
\end{align*}
When $\conf(x)$ is close to $0$, the network is not confident whether to classify $x$
as the top class $p$ or the second-best guess.
The zero set $\conf^{-1}(0)$ is the decision boundary, by definition.
Outliers of one class scattered among the points of another introduce spurious
extrema in the confidence function. By driving optimization towards the
simplified version of $\conf$, we can reduce overfitting.

Because generically $\phi(x)$ is never zero on an input point $x \in \hat{X}$,
we need an extra step to capture the topology of the decision boundary. If two
vertices $u$ and $v$, connected by an edge in the \knn graph, are assigned
two different classes by the network, then the decision boundary
passes somewhere
between them. In this case, we remove the edge $(u,v)$ from the graph.
This pruning results in multiple connected components, at least one
per class. We compute the merge tree --- forest in this case --- of the
confidence function on the pruned graph, with respect to the super-level sets,
i.e., tracking persistence of the maxima.
Because confidence function is never negative, we restrict the
infinite branches in the merge tree to die at $0$. This obviates
special treatment of separate connected components in the graph: if one of them
produces a low-persistence merge tree, we simplify it by setting the values of
all of its vertices to 0.


\section{Comparison with Diagram Simplification}
\label{sec:comparison}


Earlier work on applying topological regularization to neural
networks~\cite{topological-regularizer, topology-layer} relied on
backpropagation through persistence diagrams.
For piecewise-linear functions on a graph, each point in the 0-dimensional
persistence diagram corresponds to a pair of vertices, $(b_i, d_i) = (f(x),
f(y))$. 
If one adds a regularization term
of the form $\sum (d_i - b_i)^2$, where the sum is taken over all
points $(b_i, d_i)$ with persistence less than $\ee$,
then one can back-propagate the gradient to the function values
and then to the model parameters, i.e., the weights of the network.
We call this loss the \emph{diagram loss}, and the loss
proposed in the previous section, the \emph{PSO loss}.

The first disadvantage of the diagram loss is that only critical points generate
pairs in the persistence diagram. Accordingly, most input points are not used
and receive no information during the backpropagation.
To illustrate this, we take $f \colon \Rsp^2 \to \Rsp$ to be the sum of $4$ Gaussians
and evaluate
$f$ on the uniform grid over unit square $[0, 1] \times [0,1]$ with $10,000$ vertices.
\cref{fig:values-original} illustrates the plot of $f$.
We pick $\ee$ so that the two lower persistence points in the diagram of $f$
(corresponding to the two Gaussians with lower peaks)
are simplified, and take 50 steps of gradient descent
using the PSO loss and the diagram loss directly
on values of $f$ at each vertex. The simplified
functions appear in \cref{fig:simplified-values-pso,fig:simplified-values-dgm}, respectively.

\cref{fig:vineyard-values-pso,fig:vineyard-values-dgm} show the vineyards of the
two optimization processes.
In both vineyards, we show the original persistence values in black, the desired
values in red, and the values at each step of the optimization in green.
With PSO loss, this is an unconstrained convex problem, so the optimizer quickly
eliminates the two noisy bumps of the function, while preserving its persistent
part.
In contrast, each step of the diagram loss changes values only at critical
points, making the optimization process much slower --- after $50$ steps
both bumps are still present.
It also requires recomputing persistence diagram after each step.
This not only makes the process slower, but also introduces additional
topological noise, evident in the vineyard.
%

\begin{figure}[ht]
    \centering
    \begin{subfigure}[t]{.19\textwidth}
        \centering
        \begin{tikzpicture}[scale=0.5]
            \begin{axis}[xlabel=step, ylabel={},ylabel shift=1 mm]
                \pgfplotstableread[col sep=semicolon]{data/vineyard-weights-pso.txt}\psodata
            \addplot[
                 scatter src=explicit symbolic,
                 scatter/classes={r={mark=*,red}, b={mark=*,black}, g={mark=*,green!60!black} },
                 scatter, only marks,
                 ]
                 table[x=x,y=y,meta=label] {\psodata};
            \addplot[mark=none, black, dashed] table[x=x, y=epsilon] {\psodata};
            \end{axis}
        \end{tikzpicture}
        \caption{}
    \end{subfigure}
    \qquad
    \begin{subfigure}[t]{.19\textwidth}
        \centering
        \begin{tikzpicture}[scale=0.5]
            \begin{axis}[xlabel=step,
                         ytick style={draw=none},
                         yticklabels={,,}
                         ylabel={},
                         ylabel shift=1 mm]
                \pgfplotstableread[col sep=semicolon]{data/vineyard-weights-dgm.txt}\dgmdata
            \addplot[
                 scatter src=explicit symbolic,
                 scatter/classes={r={mark=*,red}, b={mark=*,black}, g={mark=*,green!60!black} },
                 scatter, only marks,
                 ]
                 table[x=x,y=y,meta=label] {\dgmdata};
            \addplot[mark=none, black, dashed] table[x=x, y=epsilon] {\dgmdata};
            \end{axis}
        \end{tikzpicture}
        \caption{}
    \end{subfigure}
    \caption{Optimization of the weights.
             (a) Vineyard of simplification with PSO loss. (b) Vineyard of simplification with diagram loss.}
    \label{fig:dgm-vs-pso-weights}
\end{figure}
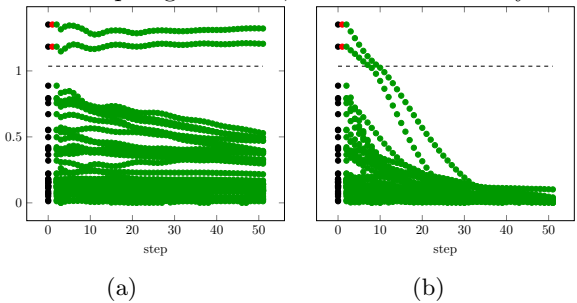

\cref{fig:dgm-vs-pso-weights} shows the effect of the two losses on a neural network.
We train a fully connected network with $5$ layers for $100$ epochs and then perform
$30$ steps of topological optimization. The key difference from the previous
example is that we do not have direct control over function values, but only
over the weights of the network.  The diagram loss provides information only for
the critical points of the function, and the optimizer ends up minimizing this
loss by pushing the whole function towards a constant: in the vineyard on the
right-hand side, all points, not just the points below $\ee$, are moving to $0$.
Since the PSO loss penalizes changes to the high-persistence parts of the
function, its optimization does not suffer from the same problem, as the
vineyard on the left-hand side shows.


It is not clear how to fix this overzealousness of the diagram loss.
The main difficulty is that the critical vertices and their pairing
change after each gradient descent step.
A naive fix would be to add a term that pushes high-persistence points
to $\infty$: $-\lambda \sum_{(d_i - b_i) > \ee} (b_i - d_i)^2$.
We have tried this approach, but it did not perform well.
Depending on weight $\lambda$, either the additional term had no influence
at all, and the function was squashed to a constant; or it dominated, and the
function exploded numerically.

A more principled solution would be to compute a matching between the
persistence diagram after each step of the topological optimization and the
target simplified diagram. The matching would translate into a loss that would
simplify the diagram, while trying to preserve the high-persistence points.
%
However, this approach has many drawbacks.
The computation of the matching, even using the fast
algorithms~\cite{geometry-helps-compare-pds}, is prohibitively expensive and
would make this procedure completely impractical.
The method itself, by construction, would only preserve the structure of the
persistence diagram, not its values at individual vertices.
Finally, changing the diagram loss function at each step of the gradient descent
may have unexpected effects on the momentum.
%


\section{Illustrative Example}
\label{sec:example}

\begin{figure*}[ht]
    \centering
 \includegraphics[width=\textwidth]{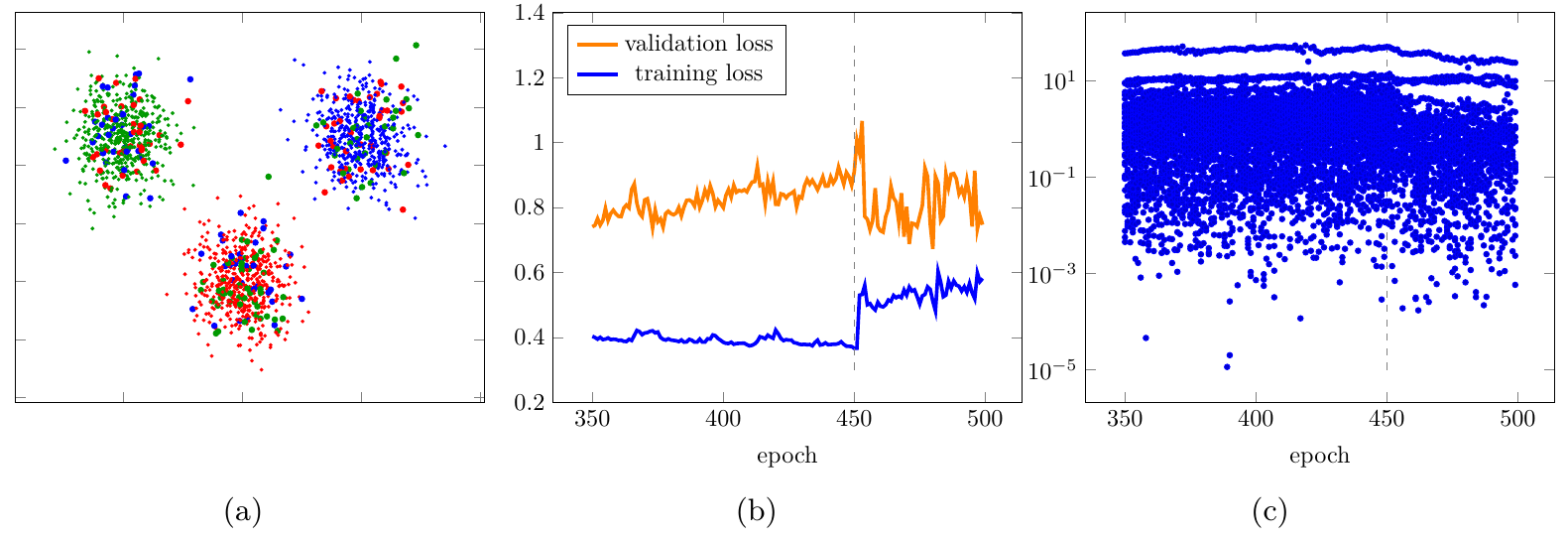}
    \caption{(a) Input data: 1,000 points sampled from each of the three
             Gaussians, representing three distinct classes, with
             20\% of the labels randomly shuffled. (b) Training and validation
             loss during the training of a neural network, restricted to the
             later epochs, where the network overfits the data. Simplification
             is applied after every epoch, following epoch 450, marked with a
             dashed line. (c) Vineyard of the confidence function during
             training; the start of the simplification is marked with a dashed
             line. The three persistent points, representing the three classes
             in the data, become prominent after the simplification.}
    \label{fig:3blob}
\end{figure*}


To illustrate how topological regularization using the PSO loss can reduce
overfitting, we consider a simple three-class dataset, shown in \cref{fig:3blob}a.
It consists of points sampled from three Gaussians, 1,000 points from each, that
represent three distinct classes.
We randomly shuffle 20\% of the labels to introduce class noise.
We train a fully-connected feedforward neural network with $5$ hidden layers of
$100$ nodes each for 500 epochs.

\cref{fig:3blob}b illustrates the training and validation losses, and
\cref{fig:3blob}c shows the persistence vineyard of the confidence
function for epochs 350 to 500.
In the beginning of this range, the network has already overfit the labels. The
growing validation loss confirms the overfitting, which is also evident in the
vineyard, where the second and third highest persistence points, which represent
the true classes in the data, are becoming indistinguishable from the noisy
points.

Starting with epoch 450, we apply ten steps of topological simplification after
every training epoch.
Because we expect each of the three classes to be a single cluster, we set $\ee$
to keep the three highest points in the persistence diagram.
This defines a PSO loss that encourages removing maxima of the confidence
function that do not correspond to the 3 predominant class clusters.

As \cref{fig:3blob}b illustrates, after turning on
simplification at epoch 450, the validation loss decreases by over $20\%$.
\cref{fig:3blob}c demonstrates the
abundance of high persistence features prior to epoch 450.  Most of these
correspond to mountains in the confidence function around noisy mislabeled points.
Turning on simplification at epoch 450 reduces the persistence of these peaks
which drives the network to match the class labels of the dominant class around
the outliers.







This toy example demonstrates how PSO simplification identifies regions of
overfitting due to class noise and reduces the confidence function near these
noisy labeled points, lowering the validation loss and increasing the accuracy
of the model after overfitting has occured.

\section{Experiments}
\label{sec:experiments}

\begin{table*}[ht]
    \centering
    \begin{tabular}{| l || l | l | l | r || l | l | l | l |}
    \hline
        & \multicolumn{3}{c|}{Regularization} & & \multicolumn{4}{c|}{Hyperparameters} \\
        Datasets        & None  & $\reg{2}$ & PSO & $\Delta$ & $k$ & $t$ & $n$ & $\sigma$ \\
    \hline
        Wine            & 0.78  & 0.77  & \best{0.76}  & 2.6\%  & 15 & 0.001  & 6  & 0.001   \\
        Iran housing    & 0.12  & 0.11  & \best{0.10}  & 16.7\% & 10 & 0.01   & 9  & 0.001    \\
        Boston          & 0.33  & 0.32  & \best{0.31}  & 6.1\%  & 20 & 0.001  & 9  & 0.001    \\
        Concrete        & 0.31  & 0.30  & \best{0.29}  & 6.4\%  & 15 & 0.01   & 3  & 0.001     \\ 
        CT slices       & 0.031 & 0.031 & \best{0.029} & 6.4\%  & 60 & 0.0001 & 1  & 0.001 \\
        Protein         & 0.64  & 0.63  & \best{0.62}  & 3.1\%  & 20 & 0.001  & 6  & 0.001  \\
    \hline
    \end{tabular}
    \caption{RMSD results on regression datasets comparing no regularization,
            $\reg{2}$ regularization of the weights, and topological
            simplification, averaged over multiple trials. The best model for
            each dataset is in bold. As topological simplification always
            results in performance improvement, the percentage of improvement
            (decrease in RMSD), from None to PSO, is also shown ($\Delta$). The
            last four columns show the hyperparameters for the best model.}
    \label{tab:uci-regression}
\end{table*}

We study the performance of persistence-sensitive optimization on six regression
problems and seven classification problems from the UCI repository
\cite{Dua2019}. To represent a variety of problem settings, the selected
datasets vary in the number of features, sample size, and number of classes. We
standardize the features by subtracting the mean and dividing by the standard
deviation. For both regression and classification, we use a dense neural network
with five hidden layers and 100 hidden nodes per layer. We use the Adam
optimizer and a learning rate of 0.001 across all experiments, including regular
training and training with topological simplification.

We compare performance of the networks trained (1) without regularization, (2)
with $\reg{2}$ regularization, (3) with topological regularization. For all
experiments, training with and without regularization were run for the same
number of total epochs.
For the $\reg{2}$ regularization, the square of the weights of the network is added
to the loss, scaled by a factor of $\lambda$, which we choose by sweeping
through a logarithmically spaced grid from $[10^{-5}, 10^1]$.  We report the
best performance across all $\lambda$s for each dataset.  For each dataset, we
run all the models at least five times with different preset random seeds and
average over all the trials.

As described in \cref{sec:method}, we set a number of hyperparameters
during the topological simplification:
\begin{itemize}[noitemsep,nolistsep]
    \item
        topological simplification is applied when validation loss increases by
        more than $t$;
    \item
        $k$ determines the number of neighbors in the \knn graph used to
        approximate the domain of the function;
    \item
        $n$ is the number of additional points we sample, for each input point,
        before building the \knn graph;
    \item
        the points are drawn from a Gaussian with variance $\sigma$, ranging from 0.001 to 0.2.
\end{itemize}
Supplementary materials list extra details for the data sets, including what
hyperparameter ranges were swept during the experiments. We always set $\ee$ to
the validation loss.
%

\paragraph{Regression.}
We evaluate the performance of topological regularization on six regression
datasets.  They vary in size from hundreds (Iran housing, Boston) to thousands
(Wine, Concrete), to tens of thousands (CT slices, Protein) data points.
For the largest dataset, CT slices, we project the data onto the first ten
principle components before computing the \knn graph.
We use a 56\%-19\%-25\% training-validation-test
split, i.e., first applying a 75\%-25\% training-test split, and then further
splitting the training set 75\%-25\% into a validation set.
We evaluate the quality of the prediction using the root-mean-square-deviation,
$\sqrt{\sum (\hat{y}_i - y_i)^2 / n}$.

\cref{tab:uci-regression} presents the results of our regression experiments.
Overall, topological simplification reduces RMSD across all the datasets by an
average of 6.9\%. Sampling each point multiple times with a small amount of
perturbation improves performance. By applying simplification when validation
loss increases by more than threshold $t$, we reduce overfitting and the
resulting error.  We also see that across the $\lambda$ hyperparameter swept for
$\reg{2}$ regularization, the performance is always worse than with topological
simplification.
We note that our method is fast enough to be used on very large datasets (we
give two examples with 40,000+ points, but that's by no means the limit); previous
approaches to topological regularization (using a form of diagram
loss)~\cite{topological-regularizer} were limited to much smaller datasets
(hundreds to a thousand points).

\begin{table*}[ht]
    \centering
    \begin{tabular}{| l || c | c | c | r || c | c | c | r || l | l | r | l |}
    \hline
        & \multicolumn{4}{c||}{Crossentropy} & \multicolumn{4}{c||}{Accuracy} & \multicolumn{4}{c|}{\multirow{2}{*}{Hyperparameters}} \\
        \cline{1-9}
        & \multicolumn{3}{c|}{Regularization} & & \multicolumn{3}{c|}{Regularization} & &\multicolumn{4}{c|}{} \\
        Datasets        & None & $\reg{2}$         & PSO           & $\Delta$   & None & $\reg{2}$          & PSO             & $\Delta$ & $k$ & $t$ & $n$ & $\sigma$ \\
    \hline
    Wisconsin cancer    & 0.13 & \best{0.08}          & 0.09 & 30.8\%     & 0.97 & 0.98            & \best{0.99}  & 2.1\%    & 15  & 0.0001 & 3   & 0.2 \\
    Wine                & 0.97 & 0.96          & \best{0.93} & 4.1\%      & 0.59 & 0.60            & \best{0.65}  & 10.2\%   & 15  & 0.0001 & 0   & 0.001 \\
    Semeion             & 0.48 & 0.48          & \best{0.38} &20.8\%      & 0.87 & 0.88            & \best{0.90}  & 3.4\%    & 20  & 0.0001 & 9   & 0.2 \\
    Vertebral           & 0.39 & 0.38          & \best{0.34} & 12.8\%     & 0.82 & \best{0.84}     & \best{0.84}  & 2.4\%    & 15  & 0.0001 & 9   & 0.1 \\
    Wireless            & 0.07 & \best{0.06}   & \best{0.06} &14.3\%      & 0.97 & 0.98            & \best{0.98}  & 1.1\%    & 25  & 0.0001 & 6   & 0.2 \\
    SPECT               & 0.35 & 0.34          & \best{0.33} & 5.7\%      & 0.80 & \best{0.83}     & 0.80         & 0\%      & 10  & 0.0001 & 15  & 0.01 \\ 
    Letter recognition  & 0.27 & 0.26          & \best{0.23} & 14.8\%     & 0.92 & 0.92            & \best{0.93}  & 1.1\%    & 20  & 0.01   & 3 & 0.2 \\
    \hline
    \end{tabular}
    \caption{Cross-entropy loss and accuracy results on classification datasets
            comparing no regularization, $\reg{2}$ regularization of the
            weights, and topological simplification, averaged over multiple
            trials. The best model is in bold. As the improvement from
            topological simplification is always greater than or equal to
            training the model without regularization, the percentage of
            improvement (decrease in the case of X-E loss and increase in the
            case of accuracy), from None to PSO, is also shown ($\Delta$). The
            last four columns show the hyperparameters for the best model, with
            the lowest X-E loss.}
    \label{tab:uci-classification}
\end{table*}

\paragraph{Classification.}
We also evaluate our method on seven classification datasets. Each one has from two to 26
classes. Similar to the regression datasets, each has hundreds
(Wisconsin cancer, Vertebral, SPECT) to thousands (Wine, Semeion, Wireless) to
tens of thousands (Letter recognition) data points.
We use the same 56\%-19\%-25\% training-validation-test split.
When topological simplification is applied, we set $\ee$ to the cross-entropy
loss and simplify the confidence function $\conf$, described in \cref{sec:method}.
We evaluate the quality of our predictions by computing the cross-entropy (X-E)
loss and accuracy.

\cref{tab:uci-classification} shows the results of our classification experiments.
The X-E loss decreases when we apply topological regularization except for the Wisconsin cancer dataset, while
accuracy increases for all the datasets, except the SPECT dataset
(the smallest dataset in size). Overall, X-E loss decreases by an average of
14.8\% and accuracy increases by an average of 2.9\% across all datasets. The
table shows the hyperparameters for the model with the lowest X-E loss. In contrast
with regression, on average, more aggressive perturbation of the sampled points
results in better performance. The best model performance across all the
datasets, except letter recognition, occurs for validation loss threshold $t$
equal to 0.0001, indicating that applying simplification as soon as validation
loss increases, i.e., as soon as the model shows any sign of overfitting, helps
regularize the training.
Topological simplification is slightly less accurate than $\reg{2}$ regularization
on the SPECT dataset, equally accurate on the vertebral dataset, the same
in terms of X-E loss on the wireless dataset, and better on all other datasets.

\begin{figure*}[ht]
    \centering
    \includegraphics[width=\textwidth]{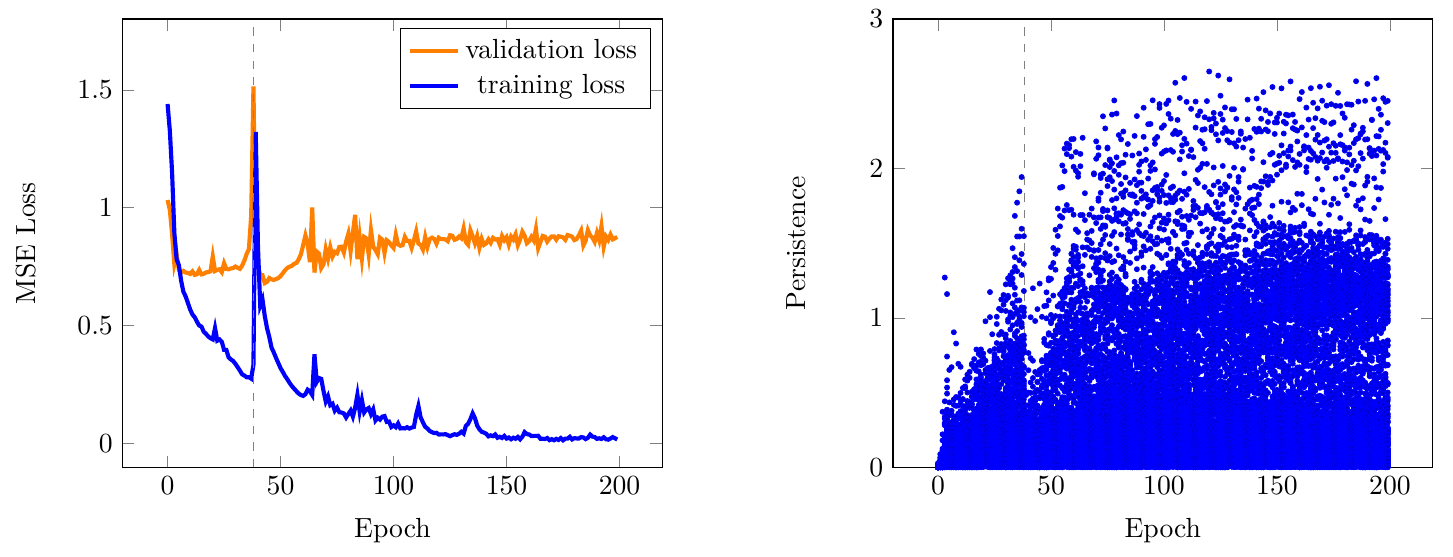}
\caption{(a) Training and validation loss curves for an experiment on the wine regression dataset.
         Performance is best at epoch 44, and simplification is applied only once, after epoch 43.
         (b) Vineyard over all epochs.}
\label{fig:wine-single-simplification}
\end{figure*}

\paragraph{Loss and vineyard.}
To better understand topological simplification, we examine the
training and validation loss curves as well as the vineyards for regression experiments
on the Wine dataset.
As \cref{fig:wine-single-simplification} illustrates, the
network quickly starts to overfit --- without simplification, within 10--15
epochs --- and the validation loss rises.
Applying simplification quickly reduces the validation loss, seemingly pushing
the system into another region of the loss landscape.
This is further seen in the vineyard, where the sharp decrease in validation
loss matches with the simplification of the persistence diagram.

\section{Conclusion}
\label{sec:conclusion}

We presented a topological regularization method that uses persistent homology,
merge trees, and persistence-sensitive simplification to minimize the number of
noisy extrema in a machine learning model. Unlike previous such methods, our
approach is faster --- requiring to compute the topological descriptor only once
per simplification phase --- as well as more robust and predictable in its effects on
the model. The key distinction of the method is its ability to prescribe
gradients on the entire domain, approximated as a \knn graph, rather than only
on the critical points. We illustrated the benefits of its use in experiments
with a number of well-known data sets.

Our work has a larger implication for the use of topological methods in machine
learning. The realization that one can back-propagate gradients through a
persistence diagram has generated considerable interest in the community,
with a number of recent works~\cite{topological-regularizer,topology-layer,differential-calculus-barcodes,HKN2019,perslay}
exploring this idea. Our results suggest that it may be better to not treat
persistence as a black box. Rather, it is a rich language that allows one to
precisely express topological constraints and priors to add to a
problem. The actual enforcement of these constraints can be accomplished via
different methods, back-propagation through the persistence diagram being but one
of them.


Building on prior work in computational topology, we describe only how to
simplify extrema, i.e., 0-dimensional persistence diagrams.  A key research
direction is how to adapt these ideas to higher dimensional persistent homology.
It is undoubtedly useful to incorporate higher-dimensional topological
constraints, such as loops or voids in the data, into optimization. Doing so
efficiently may require imposing constraints not only on the points in the
persistence diagrams, but on the entire representative cycles implied by those
points.

\vspace{-1ex}
\section{Acknowledgements}
\vspace{-1ex}

This work was supported by Laboratory Directed Research and Development (LDRD)
funding from Berkeley Lab, provided by the Director, Office of Science, of the
U.S.\ Department of Energy under Contract No.\ DE-AC02-05CH11231. A.S.K.\ was
supported by the Alvarez Fellowship in the Computational Research Division at
LBNL.

\bibliographystyle{unsrt}
\bibliography{persistence-sensitive-optimization}

\begin{thebibliography}{10}

\bibitem{convex-optimization}
Stephen~P Boyd and Lieven Vandenberghe.
\newblock {\em Convex Optimization}.
\newblock Cambridge University Press, 2004.

\bibitem{learning-with-kernels}
Bernhard Sch{\"o}lkopf, Alexander~J Smola, Francis Bach, and {Managing Director
  of the Max Planck Institute for Biological Cybernetics in Tubingen Germany
  Profe Bernhard Scholkopf}.
\newblock {\em Learning with Kernels: Support Vector Machines, Regularization,
  Optimization, and Beyond}.
\newblock MIT Press, 2002.

\bibitem{Ng2004}
Andrew~Y Ng.
\newblock Feature selection, {$L_1$} vs.\ {$L_2$} regularization, and
  rotational invariance.
\newblock In {\em Proceedings of the twenty-first international conference on
  Machine learning}, page~78, 2004.

\bibitem{topological-regularizer}
Chao Chen, Xiuyan Ni, Qinxun Bai, and Yusu Wang.
\newblock A topological regularizer for classifiers via persistent homology.
\newblock In {\em Proceedings of the International Conference on Artificial
  Intelligence and Statistics (AISTATS)}, pages 2573--2582, 2019.

\bibitem{topology-layer}
Rickard Br{\"u}el-Gabrielsson, Bradley~J Nelson, Anjan Dwaraknath, Primoz
  Skraba, Leonidas~J Guibas, and Gunnar Carlsson.
\newblock A topology layer for machine learning.
\newblock In {\em Proceedings of the International Conference on Artificial
  Intelligence and Statistics (AISTATS)}, pages 1553--1563, 2020.

\bibitem{differential-calculus-barcodes}
Jacob Leygonie, Steve Oudot, and Ulrike Tillmann.
\newblock A framework for differential calculus on persistence barcodes.
\newblock October 2019.

\bibitem{simplification-2manifolds}
Herbert Edelsbrunner, Dmitriy Morozov, and Valerio Pascucci.
\newblock Persistence-sensitive simplification functions on 2-manifolds.
\newblock In {\em Proceedings of the Annual Symposium on Computational
  Geometry}, pages 127--134. ACM, 2006.

\bibitem{simplification-linear-time}
Dominique Attali, Marc Glisse, Samuel Hornus, Francis Lazarus, and Dmitriy
  Morozov.
\newblock Persistence-sensitive simplification of functions on surfaces in
  linear time, 2009.
\newblock Manuscript. Presented at TopoInVis'09.

\bibitem{Bauer2012}
Ulrich Bauer, Carsten Lange, and Max Wardetzky.
\newblock Optimal topological simplification of discrete functions on surfaces.
\newblock {\em Discrete \& computational geometry}, 47(2):347--377, 2012.

\bibitem{HKN2019}
Christoph~D Hofer, Roland Kwitt, and Marc Niethammer.
\newblock Learning representations of persistence barcodes.
\newblock {\em Journal of machine learning research: JMLR}, 20(126):1--45,
  2019.

\bibitem{perslay}
Mathieu Carriere, Frederic Chazal, Yuichi Ike, Theo Lacombe, Martin Royer, and
  Yuhei Umeda.
\newblock {PersLay}: A neural network layer for persistence diagrams and new
  graph topological signatures.
\newblock {\em Stat}, 1050:17, 2019.

\bibitem{persistence-images}
Henry Adams, Tegan Emerson, Michael Kirby, Rachel Neville, Chris Peterson,
  Patrick Shipman, Sofya Chepushtanova, Eric Hanson, Francis Motta, and Lori
  Ziegelmeier.
\newblock Persistence images: A stable vector representation of persistent
  homology.
\newblock {\em Journal of machine learning research: JMLR}, 18(1):218--252,
  2017.

\bibitem{comp-top-book}
Herbert Edelsbrunner and John Harer.
\newblock {\em Computational topology: an introduction}.
\newblock American Mathematical Soc., 2010.

\bibitem{geometry-helps-compare-pds}
Michael Kerber, Dmitriy Morozov, and Arnur Nigmetov.
\newblock Geometry helps to compare persistence diagrams.
\newblock {\em J. Exp. Algorithmics}, 22:1.4:1--1.4:20, September 2017.

\bibitem{Dua2019}
Dheeru Dua and Casey Graff.
\newblock {UCI} machine learning repository, 2017.

\end{thebibliography}

\onecolumn
\aistatstitle{Topological Regularization via Persistence-Sensitive Optimization: \\
Supplementary Materials}

 \setcounter{section}{0}
 \setcounter{table}{0}

\section{Hyperparameter ranges for experiments}

The computation relies on the following hyperparameters:
\begin{itemize}[noitemsep,nolistsep]
    \item
        topological simplification is applied when validation loss increases by
        more than $t$;
    \item
        $k$ determines the number of neighbors in the \knn graph used to
        approximate the domain of the function;
    \item
        $n$ is the number of additional points we sample, for each input point,
        before building the \knn graph;
    \item
        the points are drawn from a Gaussian with variance $\sigma$.
\end{itemize}
The tables list the values of the hyperparameters we tried for each dataset.
In the main text, we report the model that has the best performance, highlighted in bold here.

\subsection{Regression}

\begin{table*}[h]
    \centering
    \begin{tabular}{| c | c | c | c | c |}
     \hline
     Datasets & $k$ & $t$ & $n$ & $\sigma$  \\
     \hline
     Wine & 10, \best{15}, 20 & \best{0.001}, 0.01, 0.05, 0.1, 0.5 & 0, 3, \best{6}, 9,12 & \best{0.001}, 0.01, 0.1. 0.2  \\
     Iran housing & \best{10}, 15, 20 & 0.001, \best{0.01}, 0.05, 0.1 & 0, 3, 6, \best{9}, 12 & \best{0.001}, 0.01, 0.1, 0.2 \\
     Boston & 10, 15, \best{20} & \best{0.001}, 0.01, 0.05, 0.1 & 0, 3, 6, \best{9}, 12 & \best{0.001}, 0.01, 0.1, 0.2 \\
     Concrete & 10, \best{15}, 20 & 0.001, \best{0.01}, 0.05, 0.1, 0.5 & 0, \best{3}, 6, 9, 12 & \best{0.001}, 0.01, 0.1, 0.2 \\
     CT slices & 20, 40, \best{60}, 80 & \best{0.0001}, 0.001 & 0, \best{1} & \best{0.001}, 0.01, 0.1, 0.2 \\
     Protein & \best{20}, 40, 60, 80 & \best{0.001}, 0.01, 0.1 & 0, 3, \best{6} & \best{0.001}, 0.01, 0.1, 0.2 \\
     \hline

    \end{tabular}
    \caption{Hyperparameter ranges for regression datasets.}
    \end{table*}

\subsection{Classification}

\begin{table*}[h]
    \centering
    \begin{tabular}{| c | c | c | c | c |}
     \hline
     Datasets & $k$ & $t$ & $n$ & $\sigma$  \\
     \hline
     Wisconsin cancer & 10, \best{15}, 20 & \best{0.0001}, 0.001, 0.01, 0.1 & 0, \best{3}, 6, 9,12 & 0.001, 0.01, 0.1. \best{0.2}  \\
     Wine & 10, \best{15}, 20 & \best{0.0001}, 0.001, 0.01, 0.1 & \best{0}, 3, 6, 9, 12 & \best{0.001}, 0.01, 0.1, 0.2 \\
     Semeion & 10, 15, \best{20} & \best{0.0001}, 0.001, 0.01, 0.1 & 0, 3, 6, \best{9}, 12 & 0.001, 0.01, 0.1, \best{0.2} \\
     Vertebral & 10, \best{15}, 20 & \best{0.0001}, 0.001, 0.01, 0.1  & 0, 3, 6, \best{9}, 12 & 0.001, 0.01, \best{0.1}, 0.2 \\
     Wireless & 10, 15, 20, \best{25} & \best{0.0001}, 0.001, 0.01, 0.1 & 0, 3, \best{6}, 9, 12 & 0.001, 0.01, 0.1, \best{0.2} \\
     SPECT & \best{10}, 15, 20 & \best{0.0001}, 0.001, 0.01, 0.1 & 0, 3, 6, 9, 12, \best{15} & 0.001, \best{0.01}, 0.1, 0.2 \\
     Letter recognition & 10, \best{20}, 30, 40, 50 & 0.0001, 0.001, \best{0.01} & 0, \best{3}, 6, 9, 12, 15 & 0.001, 0.01, 0.1, \best{0.2} \\
     \hline

    \end{tabular}
    \caption{Hyperparameter ranges for classification datasets.}
    \end{table*}

\end{document}